\documentclass[11pt]{article}

\usepackage[final]{acl}

\usepackage{times}
\usepackage{latexsym}

\usepackage[T1]{fontenc}

\usepackage[utf8]{inputenc}

\usepackage{microtype}

\usepackage[table]{xcolor}  

\usepackage{inconsolata}
\usepackage[most]{tcolorbox}

\usepackage{graphicx}
\usepackage{amsmath}
\usepackage{enumitem}
\usepackage{booktabs}
\usepackage{multirow}
\usepackage{algorithm}
\usepackage{algpseudocode}
\usepackage{graphicx}
\usepackage{subcaption}
\newcommand{\annotate}[3]{%
    #1\raisebox{-0.5ex}{\scriptsize\textcolor{#2}{#3}}%
}

\usepackage{array}
\newcolumntype{C}[1]{>{\centering\arraybackslash}p{#1}}

\newtcolorbox{promptbox}[1][]{
  colback=gray!5!white,
  colframe=black!75!black,
  fonttitle=\bfseries,
  colbacktitle=gray!85!black,
  title=#1,
  width=\textwidth,           
  boxrule=0.3pt,
  arc=2pt,
  left=5pt, right=5pt, top=4pt, bottom=4pt,
  before skip=8pt, after skip=8pt,
  breakable,
  enhanced jigsaw
}

\title{When to Trust Tools? Adaptive Tool Trust Calibration For Tool-Integrated Math Reasoning}

\author{
  Ruotao Xu$^{1}$,
  Yixin Ji$^{1}$,
  Yu Luo$^{2}$,
  Jinpeng Li$^{2}$,
  Dong Li$^{2}$, \\
  \textbf{Peifeng Li}$^{1}$,
  \textbf{Juntao Li}$^{1}$\thanks{\; Corresponding author.}, 
  \textbf{Min Zhang}$^{3, 1}$ \\
  $^{1}$School of Computer Science and Technology, Soochow University\\
  $^{2}$Department of Foundation Model, 2012 Labs, Huawei \\
  $^{3}$Harbin Institute of Technology, Shenzhen (HITSZ)\\
  \texttt{\{xuruotao007, jiyixin169\}@gmail.com} \hspace{1cm}
  \texttt{\{ljt,minzhang\}@suda.edu.cn}
}

\begin{document}
\maketitle
\begin{abstract}
Large reasoning models (LRMs) have achieved strong performance enhancement through scaling test time computation, but due to the inherent limitations of the underlying language models, they still have shortcomings in tasks that require precise computation and extensive knowledge reserves. 
Tool-Integrated Reasoning (TIR)  has emerged as a promising paradigm that incorporates tool call and execution within the reasoning trajectory.
Although recent works have released some powerful open-source TIR models, our analysis reveals that these models still suffer from critical deficiencies.
We find that when the reasoning of the model conflicts with the tool results, the model tends to believe in its own reasoning. And there are cases where the tool results are correct but are ignored by the model, resulting in incorrect answers, which we define as ``Tool Ignored''. This indicates that the model does not know when to trust or ignore the tool.
To overcome these limitations, We introduce Adaptive Tool Trust Calibration (ATTC), a novel framework that guides the model to adaptively choose to trust or ignore the tool results based on the confidence score of generated code blocks. The experimental results from various open-source TIR models of different sizes and across multiple datasets demonstrate that ATTC effectively reduces the ``Tool Ignored'' issue, resulting in a performance increase of 4.1\% to 7.5\%. Our codes are available at \url{https://github.com/00Dreamer00/ATTC}.
\end{abstract}

\section{Introduction}
\begin{figure}[t]
    \centering
    \includegraphics[width=\columnwidth]{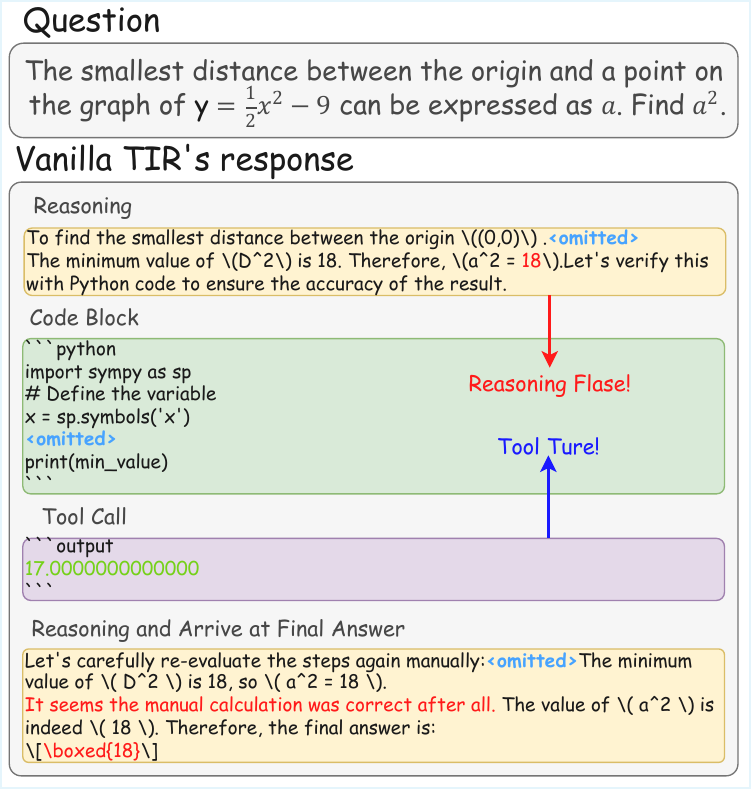}
    \caption{Case of the ``Tool Ignored'' phenomenon. The model arrive at an incorrect answer of 18 through reasoning, while the tool provide the correct result of 17. The model ignore the tool and give an incorrect answer.}
    \label{fig:case1}
\end{figure}

The rapid evolution of Large Reasoning Models (LRMs)~\citep{ openai2024openaio1card, DeepSeekAI2025DeepSeekR1IR, yang2025qwen3technicalreport, Team2025KimiKS, comanici2025gemini25pushingfrontier}  represents a transformative milestone in the history of Large Language Models (LLMs). 
By scaling test-time computation~\citep{ji2025surveytesttimecomputeintuitive}, these models have achieved substantial performance gains in handling challenging reasoning tasks.
Unlike conventional LLMs that typically generate responses in a direct, LRMs engage in long Chain-of-Thought (CoT) reasoning before generating an answer. This shift toward a systematic deliberation process allows the model to refine its logic and arrive at more robust final outputs.

Nevertheless, LRMs are still constrained by the inherent limitations of the underlying language models~\citep{zhao2025surveylargelanguagemodels, yue2025doesreinforcementlearningreally}, most notably in areas requiring precise numerical computation and comprehensive knowledge coverage. To mitigate these limitations, Tool-Integrated Reasoning (TIR)~\citep{gou2024toratoolintegratedreasoningagent, wang2023mathcoderseamlesscodeintegration, liao2024mariomathreasoningcode} has emerged as a promising paradigm that incorporates tool call and execution within the reasoning trajectory. By incorporating external tools such as code executors and search engines, TIR empowers models to transcend the performance bottlenecks of purely reasoning. 

Early works in  TIR primarily rely on prompt engineering~\citep{wang2025selfdcreasonactself, yuan2024craftcustomizingllmscreating, qian2024investigateconsolidateexploitgeneralstrategyintertask,chen2023programthoughtspromptingdisentangling, yang2024bufferthoughtsthoughtaugmentedreasoning} to guide LLMs in tool call. However, these approaches are heavily dependent on meticulously crafted prompts, which limit their scalability and generalizability. 
Some later works use supervised finetuning ~\citep{chen2025advancingtoolaugmentedlargelanguage, qian2025smartselfawareagenttool, yang2024qwen25mathtechnicalreportmathematical, yao2023reactsynergizingreasoningacting, wang2023mathcoderseamlesscodeintegration} to internalize the behavior pattern of the model actively calling tools in reasoning by training models on specialized datasets enriched with tool call demonstrations.
Nevertheless, SFT-based methodologies exhibit inherent limitations, as they constrain models to strictly adhere to the tool usage patterns present in the training data distribution, these models often fail to develop adaptive strategies.
To address these issues, several recent works focus on applying reinforcement learning~\citep{feng2025retoolreinforcementlearningstrategic, li2025torlscalingtoolintegratedrl, jiang2025verltoolholisticagenticreinforcement, bai2025effectivecodeintegratedreasoning, xue2025simpletirendtoendreinforcementlearning}  to improve the tool use ability of models. These works enable the model to have more flexible strategies when calling tools based on the complexity of the task.

In this study, we specifically focus on the application of TIR in scenarios where code executors are utilized as the tool. Although existing works have enabled the models to perform TIR, our analysis reveals that existing open-source TIR models still suffer from critical deficiencies. Most notably, there remains a persistent struggle to achieve an optimal balance between external tool result and reasoning. 
By analyzing the reasoning trajectories of TIR models, we observe a widespread contradiction between the models' reasoning and the results provided by external tools in false cases. When such conflicts arise, models often lack a robust mechanism to reconcile the divergent information, frequently opting to ignore the tool's output. 
As shown in Figure~\ref{fig:case1}, the model fails to arrive at the correct answer specifically because it ignores a valid tool result,  we define this phenomenon as ``Tool Ignored''.
This behavior indicates that current TIR models struggle to accurately discern when to trust or dismiss tool result, leading to redundant reasoning paths and erroneous conclusions.

To overcome these limitations, we propose Adaptive Tool Trust Calibration (ATTC), a novel framework that guides the model to adaptively choose to trust or ignore the tool results based on the confidence score of generated code blocks.
When a model calling tool is detected, ATTC will score the code blocks generated by the model based on a specific confidence scoring formula. If the confidence score is greater than the empirically determined threshold, ATTC will guide the model to trust the tool results, otherwise ATTC will guide the model to rethink.
We conduct extensive experiments across multiple open-source TIR models, and the experimental results decisively demonstrate that ATTC effectively reduces the ``Tool Ignored'' issue, resulting in a performance increase of 4.1\% to 7.5\%.

Overall, our contributions are as follows:
\begin{itemize} [leftmargin=*]
\setlength{\itemsep}{0pt}
\setlength{\parskip}{0pt}
    \item We find that the TIR model does not know when to trust the results of the tool and define the ``Tool Ignored'' issue.

    \item We propose a novel framework ATTC for tool-integrated reasoning that guides the model to adaptively choose to trust or ignore the tool results based on the confidence score of generated code blocks.

    \item Extensive experiments conducted on open-source TIR models of various sizes demonstrate that ATTC improves the performance of the models.

\end{itemize}

\section{Phenomenon Analysis}
\label{sec:phenomenon}

\subsection{Contradictions of Reasoning and Tools}
After careful observation of many trajectories of Tool-Integrated Reasoning, we observe that the conclusions produced by the model’s reasoning are not always consistent with the outputs returned by external tools, in fact conflicts between the two arise frequently.
To determine the exact proportion of such conflicting instances and to characterize their resolution, we conduct an LLM-based audit of more than 32k cases followed by a detailed quantitative evaluation.
Specifically, we measure the prevalence of conflicts separately in true and false cases. In cases of conflict, we further analyze whether the model tends to rely on its own reasoning or defer to tool outputs.
The prompts used to guide this analysis are provided in the Appendix~\ref{sec:appendixB}.

Figure~\ref{fig:trust_or_not} shows that a significant proportion, between 40\% and 60\%, of the false cases display a contradiction between the model's reasoning and the output from the external tool. In over half of the conflicting scenarios, the model exhibits a strong tendency to trust its own reasoning over the results produced by the tool. This pattern reveals a significant limitation in the model's meta-cognition, specifically its inability to effectively determine when to trust the output returned by the tool compared to its own reasoning results.

\begin{figure}[t]
    \centering
    \includegraphics[width=\columnwidth]{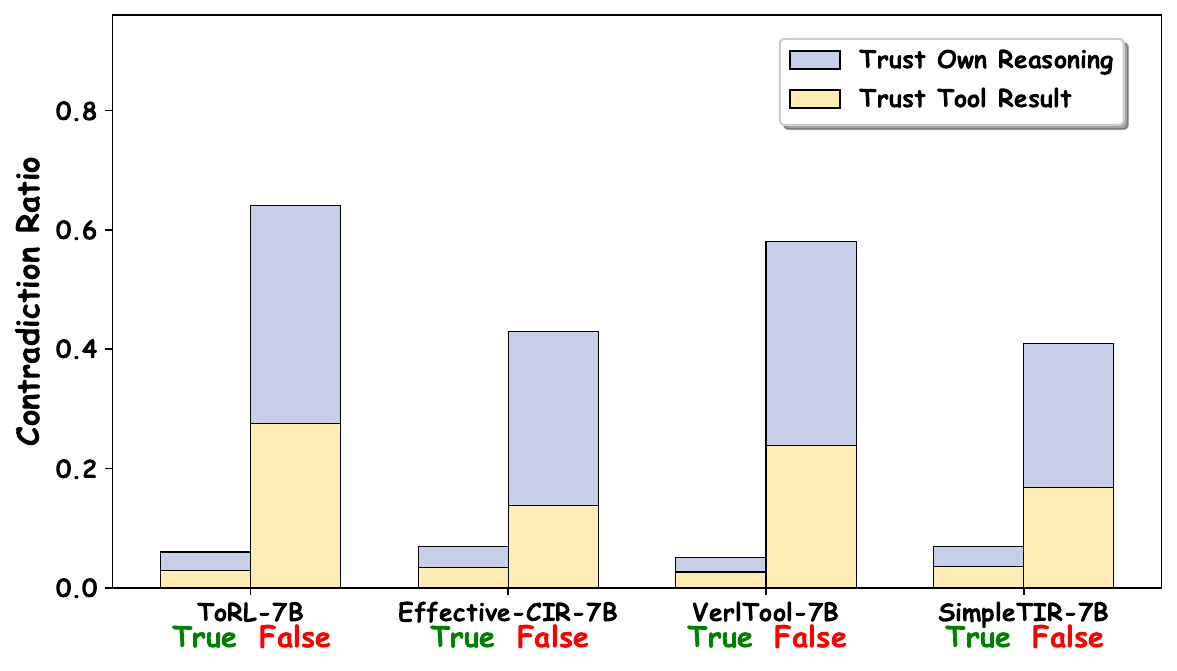}
    \caption{The proportion of contradictions between model reasoning and tool results in true and false cases. The color proportions in the column indicate the proportion that the model chooses to believe in its own reasoning or tool results.}
    \label{fig:trust_or_not}
\end{figure}

\subsection{Tool Ignored}
By examining a large set of false cases exhibiting reasoning tool contradictions, we identify a counterintuitive failure mode: when a conflict arises, the tool output is correct, yet the model ignores it and instead adheres to its own reasoning, producing an incorrect answer. We name this failure mode as ``Tool Ignored''. This phenomenon indicates a systematic preference for self-generated reasoning over externally provided evidence, which prevents the model from fully leveraging tool augmentation and ultimately degrades task accuracy.
A specific case is shown in  Figure~\ref{fig:case1}.

\begin{figure}[t]
    \centering
    \includegraphics[width=\columnwidth]{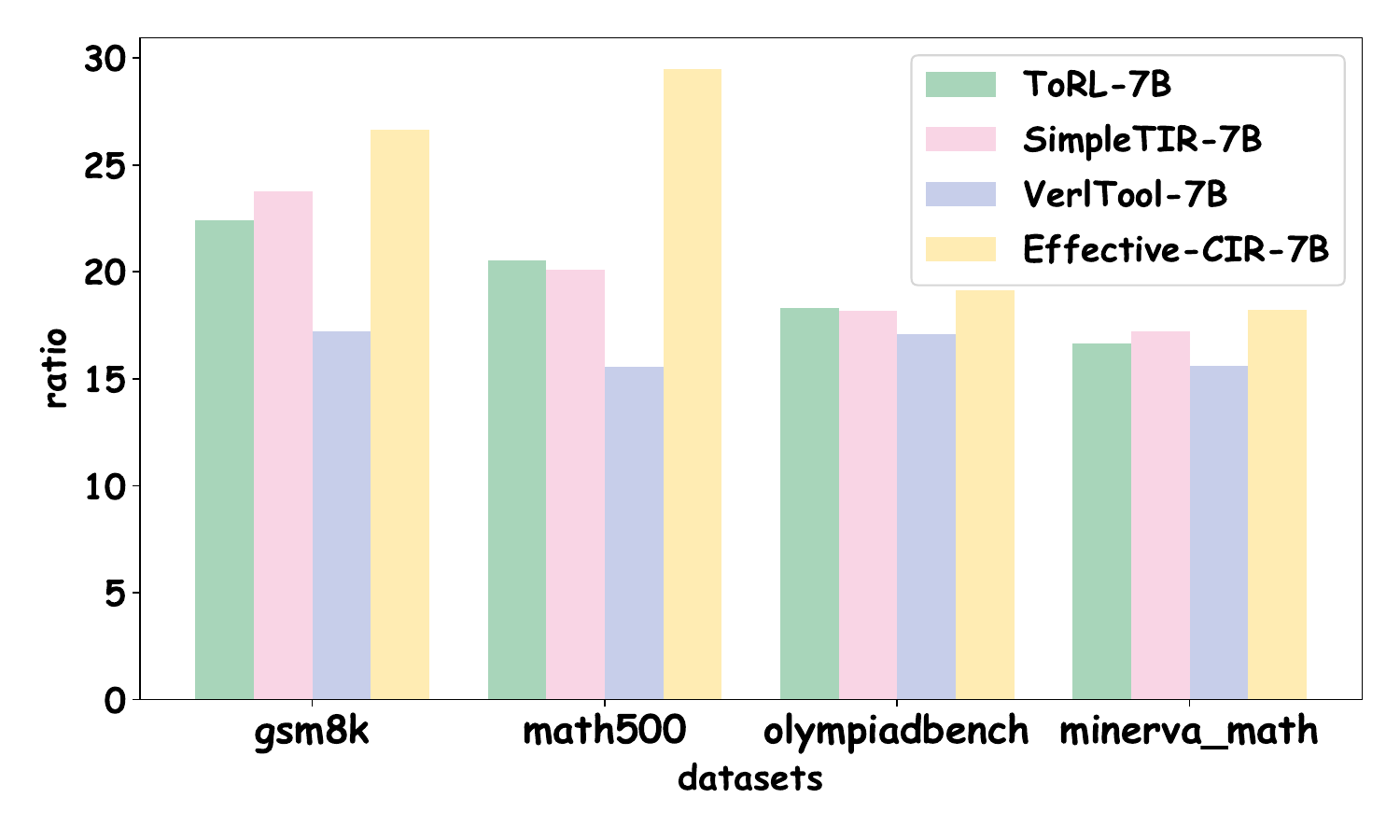}
    \caption{Proportion of ``Tool Ignored'' phenomenon of different TIR models on four datasets.}
    \label{fig:overconfidence}
\end{figure}

To assess the prevalence of this failure mode, we conduct a systematic analysis of false cases across four distinct models and four challenging datasets. The results summarized in Figure~\ref{fig:overconfidence} show that in every model–dataset combination, ``Tool Ignored'' accounts for at least 15\% of errors. This phenomenon undermines both accuracy and computational efficiency. When a verifiably correct tool output is ignored, the model often tends to generate redundant reasoning steps or tool calls, yielding an incorrect prediction while incurring unnecessary computational cost.
It is also important to avoid blindly accepting the results generated by tools. Therefore, the key challenge is to endow the model with meta-cognitive calibration to decide when to trust an external tool and when to rely on its reasoning.

\section{Methodology}
\begin{figure*}[t]
    \centering
    \includegraphics[width=\textwidth]{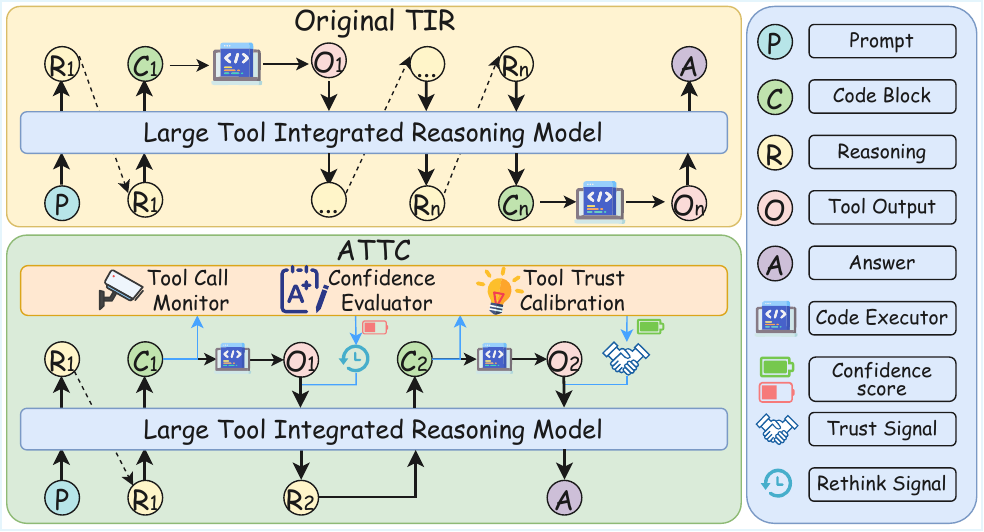}
    \caption{An overview of the Adaptive Tool Trust Calibration (ATTC) method.}
    \label{fig:method}
\end{figure*}

\newcommand{\ncolumn}{7}
\begin{table*}[t]
    \centering
    \footnotesize
    \setlength{\tabcolsep}{6.5pt}     
    \renewcommand{\arraystretch}{1.15} 
    \vspace{0.2cm}
    \begin{tabular}{lcccccl}
        \toprule
        \textbf{Model} &
        \textbf{MATH 500} &
        \textbf{Minerva Math} &
        \textbf{Olympiad} & \textbf{AIME24} & \textbf{AMC23} &
        \textbf{Avg} \\
        \midrule 
        \multicolumn{\ncolumn}{c}{\itshape Models based on Qwen2.5-7B} \\
        \midrule
        ToRL-7B                &  82.2 & 33.5 & 49.9 & 43.3 & 65.0 & 54.8 \\
        \rowcolor{blue!10} 
        \hspace{1em} +ATTC    &  84.8 & 43.8 & 52.4 & 46.7 & 72.5 &     {{\textbf{\annotate{60.0}{green!70!black}{+5.2}}}}  \\
        Effective TIR-7B       &  82.8 & 30.5 & 51.9 & 42.3 & 70.0 & 55.5 \\
        \rowcolor{blue!10} 
        \hspace{1em} +ATTC   &  85.8 & 42.3 & 53.5 & 46.7 & 77.5 & {{\textbf{\annotate{61.2}{green!70!black}{+5.7}}}} \\
        VerlTool-7B          &  82.0 & 31.6 & 49.8 & 40.0 & 67.5 & 54.2 \\
        \rowcolor{blue!10} 
        \hspace{1em} +ATTC     &  83.4 & 44.1 & 50.5 & 43.3 & 70.0 & {{\textbf{\annotate{58.3}{green!70!black}{+4.1}}}} \\
        SimpleTIR-7B          &  82.1 & 30.1 & 47.4 & 46.7 & 75.0 & 56.3 \\
        \rowcolor{blue!10} 
        \hspace{1em} +ATTC     &  83.2 & 46.7 & 49.2 & 50.0 & 77.5 & {{\textbf{\annotate{61.3}{green!70!black}{+5.0}}}} \\

        \midrule
        \multicolumn{\ncolumn}{c}{\itshape Models based on Qwen2.5-32B} \\        
        \midrule
        ReTool-32B              &  84.6 & 30.5 & 60.1 & 53.3 & 80.0 & 61.7 \\
        \rowcolor{blue!10} 
        \hspace{1em} +ATTC     &  87.4 & 36.8 & 62.5 & 66.7 & 92.5 & {{\textbf{\annotate{69.2}{green!70!black}{+7.5}}}} \\
        SimpleTIR-32B         &  85.2 & 33.8 & 53.8 & 50.0 & 80.0 & 60.6 \\
        \rowcolor{blue!10} 
        \hspace{1em} +ATTC   &  88.2 & 36.8 & 56.9 & 56.7 & 85 &{{\textbf{\annotate{64.7}{green!70!black}{+4.1}}}} \\
        \midrule
        \multicolumn{\ncolumn}{c}{\itshape Models based on Qwen3-4B} \\
        \midrule
        ReTool-4B        &  57.0 & 16.2 & 27.7 & 16.7 & 42.5 & 32.0 \\
        \rowcolor{blue!10} 
        \hspace{1em} +ATTC     &  61.8 & 23.9 & 32 & 16.7 & 52.5 & {{\textbf{\annotate{37.4}{green!70!black}{+5.4}}}} \\
        DemyAgent-4B         &  71.4 & 17.3 & 51.9 & 40.0 & 75.0 & 51.1 \\
        \rowcolor{blue!10} 
        \hspace{1em} +ATTC     &  79.4 & 22.4 & 53.5 & 43.3 & 77.5 & {{\textbf{\annotate{55.2}{green!70!black}{+4.1}}}} \\
        \bottomrule
    \end{tabular}
    \caption{\label{tab:main_results}
    Pass@1 performance of the proposed ATTC method across various tool integrated reasoning models on various math benchmarks. }
\end{table*}

\subsection{Preliminaries}
In this work, we consider a Tool-Integrated Reasoning (TIR) setting where a language model interacts with an external Python execution environment during test-time reasoning. 
Under this particular paradigm, code generation is selectively and autonomously triggered by the model during its reasoning process. Crucially, the model is trained to learn how to effectively leverage this generated code to assist, augment, and validate its reasoning capabilities.



Formally, the TIR model maintains a reasoning trajectory $\mathcal{T}^{(t)}$ at iteration $t$ as:
\begin{equation}
\label{eq:alt_tir1}
\mathcal{T}^{(t)} = \bigl\{ (r^{(1)}, c^{(1)}, o^{(1)}), \dots, (r^{(t)}, c^{(t)}, o^{(t)}) \bigr\}
\end{equation}
where $r^{(t)}$ denotes the natural language reasoning,
$c^{(t)}$ represents the generated executable code,
and $o^{(t)}$ is the execution result returned by the external environment.
The iterative generation process follows: 
\begin{equation}
\label{eq:alt_tir2}
(r^{(t)}, c^{(t)}) \sim M_{\mathrm{tir}}\!\left( Q, \mathcal{T}^{(t-1)} \right)
\end{equation}

\begin{equation}
\label{eq:alt_tir3}
o^{(t)} = \mathcal{E}\!\left( c^{(t)} \right)
\end{equation}

\begin{equation}
\label{eq:alt_tir4}
\mathcal{T}^{(t)} = \mathcal{T}^{(t-1)} \cup \bigl\{ (r^{(t)}, c^{(t)}, o^{(t)}) \bigr\}
\end{equation}
Given the input prompt $Q$ and the accumulated trajectory $\mathcal{T}^{(t-1)}$,
the TIR model $M_{\mathrm{tir}}$ continues to generate.
The generated code is then executed by an external code execution environment
$\mathcal{E}$ to obtain the corresponding output.
This iterative process continues until a termination condition is satisfied,
at which point the model produces the final answer.

To address the ``Tool Ignored'' phenomenon discussed in Section~\ref{sec:phenomenon},  we introduce the Adaptive Tool Trust Calibration (ATTC) method to 
guide the TIR model on whether to trust or ignore the tool in the next section.

\subsection{Adaptive Tool Trust Calibration}

The main idea behind ATTC is that the model's confidence in its generated code block indicates how sufficient its prerequisite reasoning is before making a tool call. We observe that code blocks resulting from incomplete or flawed reasoning processes tend to exhibit markedly lower confidence levels. Conversely, when the preceding thought process is comprehensive and logically sound, the model generates code with a significantly higher degree of certainty.
Specific quantitative experiments can be found in Section~\ref{sec:Analysis}.
This pattern suggests that the TIR model implicitly recognizes the trustworthiness of the tool's potential output but lacks the explicit mechanism to utilize this awareness in its ongoing reasoning, frequently leading to erroneous trust or ignorance of tool results. ATTC is designed to bridge this gap by converting this implicit awareness into an explicit, actionable control signal.

As shown in Figure~\ref{fig:method}, ATTC involves three designs to determine whether to trust tool: tool call monitor, confidence evaluator, tool trust recalibrate.

\paragraph{Tool Call Monitor}
Within the ATTC framework, the tool call monitor module is responsible for actively supervising the model's generation stream. Due to the  generation paradigm of TIR model, we use a rule-based monitor method. Specifically, the tool call monitor detects a tool call and subsequently suspends the model's generation flow when dedicated markers, such as `` \verb|```|output'' or ``<tool\_result>'' are encountered.
Simultaneously, we utilize the module to extract the code block generated by the model, which is usually enclosed within specific delimiters such as ``\verb|```|python \ \verb|```|'' or ``<python> \ </python>'', for use in downstream system processes. 
The code block for iteration $t$ is as follows:
$c^{(t)}$ = [$c^{(t)}_{0}$, $c^{(t)}_{1}$, $c^{(t)}_{2}$, ..., $c^{(t)}_{n}$], where $c^{(t)}_{i}$ denotes a token in the code block.

\paragraph{Confidence Evaluator}
The confidence evaluator is designed to compute the confidence of the model-generated code block. Specifically, it defines the confidence of an individual token as the maximum predicted probability assigned to that token by the model.
The i-th token $c^{(t)}_{i}$ in the code block corresponds to the k-th token generated by $M_{tir}$ at iteration $t$.
To derive a single measure for the entire code block, the overall confidence score is then computed as the geometric mean of these token-level confidence scores across all constituent tokens, the confidence $\mathcal{C}$ at iteration $t$ is  calculated as follows:
\begin{equation}
    \label{eq:TIR5}
    p(c^{(t)}_{i}) = softmax(H_{\mathrm{tir}}\!\left( Q, \mathcal{T}^{(t-1)} \right))_{k}
\end{equation}

\begin{equation}
    \label{eq:TIR6}
    \mathcal{C} = \left( \prod_{i=1}^{n} \max p(c^{(t)}_{i}) \right)^{1/n}
\end{equation}
where $H_{tir}$ is the language model head at the final layer of the TIR model $M_{tir}$. 
The calculation of $\mathcal{C}$ utilizes the
geometric mean, a choice driven by its sensitivity to low probability values, effectively reflecting the impact of tokens with lower confidence levels. In addition, geometric mean naturally has scale invariance, which can avoid bias caused by sequence length or probability scale changes, and is more suitable as an indicator to measure the overall confidence level of the code block.

\paragraph{Tool Trust Recalibration}
Finally, the comparison between the confidence score $\mathcal{C}$ and the empirically determined threshold $\lambda$ governs whether the TIR model should trust the tool result.
If $\mathcal{C} \ge \lambda$, the current tool result is deemed trustworthy, and a specific trust control signal is injected. This signal explicitly directs the model to accept the tool's output and give the answer based on that result. 
Conversely, if $\mathcal{C} < \lambda$, the model is not instructed to completely disregard the tool output. Instead, a rethink control signal is injected, which explicitly guides the model to utilize the tool result to critically reflect upon its prior reasoning and generated code, thus necessitating a full reconsideration of the entire inferential process. For detailed prompts, please refer to Appendix~\ref{sec:appendixD}.

\section{Experiments}

\subsection{Experimental Setup}
\paragraph{Benchmarks and Metrics}
To comprehensively evaluate the tool integrated reasoning capabilities of LLMs, We evaluate on multiple mathematical benchmarks: 
MATH-500~\citep{hendrycks2021measuringmathematicalproblemsolving}, Minerva~\citep{lewkowycz2022solvingquantitativereasoningproblems}, Olympiad~\citep{he2024olympiadbenchchallengingbenchmarkpromoting}, AIME24 and AMC23.
The primary evaluation metric is Accuracy, which measures the proportion of correct final answers, calculated as the average pass@1 score. In addition, we employ two auxiliary metrics, Token Count and Time Use, representing the average number of tokens generated per sample and the average inference time per sample, respectively. They are used as indicators of the efficiency of tool-integrated reasoning.


\begin{table*}[t]
    \centering
    \footnotesize
    \setlength{\tabcolsep}{4.5pt}     
    \renewcommand{\arraystretch}{1.15} 
    \vspace{0.2cm}
    \begin{tabular}{lcccccccccccl}
        \toprule
        
 \multirow{2}{*}{\textbf{Model}} &
 \multicolumn{2}{c}{\textbf{MATH 500}} & 
 \multicolumn{2}{c}{\textbf{Minerva}} &
 \multicolumn{2}{c}{\textbf{Olympiad}} &
 \multicolumn{2}{c}{\textbf{AIME24}} & 
 \multicolumn{2}{c}{\textbf{AMC23}} & 
 \multicolumn{2}{c}{\textbf{Avg}}\\
    & {Tok$\downarrow$} & {Time$\downarrow$}
    & {Tok$\downarrow$} & {Time$\downarrow$}
    & {Tok$\downarrow$} & {Time$\downarrow$}
    & {Tok$\downarrow$} & {Time$\downarrow$}
    & {Tok$\downarrow$} & {Time$\downarrow$
    }& {Tok$\downarrow$} & {Time$\downarrow$}\\

        \midrule 
        \multicolumn{13}{c}{\itshape Models based on Qwen2.5-7B} \\
        \midrule
        
        
        EffectiveTIR-7B      & 1195 & 241 & 838 & 120 & 1203 & 318 &  1638 & 129 & 1503 & 78 & 1275 & 177 \\
        \rowcolor{blue!10} 
        \hspace{1em} +ATTC  & 833 & 177 & 901 & 119 & 1283 &  316 & 1636 & 126 & 1398 & 71 & {{\textbf{\annotate{1210}{green!70!black}{-5\%}}}} &  {{\textbf{\annotate{161}{green!70!black}{-9\%}}}} \\
        
        SimpleTIR-7B          & 1538 & 389 & 2683 & 373 & 2774 & 624 & 4444 & 278 & 2737 & 241 & 2835 & 381\\
        \rowcolor{blue!10} 
        \hspace{1em} +ATTC    & 1316 & 323 & 2534 & 328 & 2898 & 653 & 3723 & 206 & 2428 & 165 & {{\textbf{\annotate{2579}{green!70!black}{-9\%}}}} &  {{\textbf{\annotate{335}{green!70!black}{-12\%}}}} \\

        \midrule
        \multicolumn{13}{c}{\itshape Models based on Qwen2.5-32B} \\        
        \midrule
        ReTool-32B          & 1476 & 469 & 1927 & 422 & 2316 & 804 &  3080 & 265 & 2126 & 200 & 2185 & 432\\
        \rowcolor{blue!10} 
        \hspace{1em} +ATTC   & 1429 & 406 & 1800 & 360 & 2042 &  650 & 2705 & 259 & 1878 & 184 & {{\textbf{\annotate{1970}{green!70!black}{-10\%}}}} &  {{\textbf{\annotate{371}{green!70!black}{-14\%}}}} \\
        SimpleTIR-32B      & 1377 & 425 & 1849 & 368 & 2430 & 636 &  2865 & 281 & 1770 & 201 & 2058 & 382\\
        \rowcolor{blue!10} 
        \hspace{1em} +ATTC    & 1204 & 393 & 1472 & 320 & 2124 & 666 & 2865 & 236 & 1520 & 186 & {{\textbf{\annotate{1837}{green!70!black}{-11\%}}}} &  {{\textbf{\annotate{360}{green!70!black}{-6\%}}}} \\
        
        
        \bottomrule
    \end{tabular}
    \caption{\label{tab:efficient}
    Efficiency performance of the proposed ATTC method across various tool integrated reasoning models on various math benchmarks. "Tok" denotes the average token count for a single question. "Time" denotes the average single reasoning time of each benchmark, and the unit is seconds. $\downarrow$ indicates that lower values are better.}

\end{table*}

\paragraph{Models}
We apply our method to a series of open-source TIR models trained through reinforcement learning. Some of them are based on Qwen2.5-7B, such as ToRL-7B~\citep{li2025torlscalingtoolintegratedrl}, VerlTool-7B~\citep{jiang2025verltoolholisticagenticreinforcement},  Effective TIR-7B~\citep{bai2025effectivecodeintegratedreasoning} and SimpleTIR-7B~\citep{xue2025simpletirendtoendreinforcementlearning}. There are also some based on Qwen2.5-32B and Qwen3-4B, such as ReTool-32B~\citep{feng2025retoolreinforcementlearningstrategic}, SimpleTIR-32B, DemyAgent-4B~\citep{yu2025demystifyingreinforcementlearningagentic} and so on.

\paragraph{Implementation Details}
All experiments are 
conducted using the vLLM framework.
For the decoding strategy, we align the temperature and top-p settings with the optimal settings provided in the corresponding model's paper. Please refer to the Appendix~\ref{sec:appendixC} for specific settings.
We conduct experiments with 3 random seeds and report the average pass@1 results.

\subsection{Experimental Results}
\paragraph{Effectiveness Experiment}
The results in Table~\ref{tab:main_results} show that ATTC consistently outperforms the vanilla  tool-integrated reasoning across three model sizes and five benchmark datasets.
As evidenced by the experimental results, the performance enhancements delivered by ATTC are comprehensive and consistent. ATTC enhances model performance by an average of 4.1\% to 7.5\% across five different datasets.
Across five benchmarks of varying difficulty, ATTC consistently demonstrates significant performance improvements, indicating that its effectiveness is independent of dataset complexity. As detailed in Section~\ref{sec:phenomenon}, TIR models systematically exhibit the ``Tool Ignored'' failure mode and often lack calibrated criteria for when to trust external tools. ATTC addresses this limitation by instructing the model on whether to trust tool outputs or the model’s reasoning, thereby improving accuracy across diverse tasks. ATTC clearly offers advantages across three different model scales, showing its effectiveness regardless of model size. Its benefits are not dependent on the model's reasoning capabilities or internal representations. Instead, they stem from optimizing the interaction between the model and the tool, as well as improving the utilization of the tool. These results indicate that ATTC serves as a general, training-independent framework for effective tool use.
The results in Table~\ref{tab:ceiling_ability} show that ATTC consistently outperforms vanilla tool-integrated reasoning models on AIME 2025 in terms of both Pass@1 and Pass@32. The experimental results prove that ATTC is effective on truly unseen, high-difficulty reasoning problems and improves the ceiling of the ability of the reasoning model.

\begin{table}[t]
    \centering
    \footnotesize
    \setlength{\tabcolsep}{3pt}     
    \renewcommand{\arraystretch}{1.15} 
    \vspace{0.2cm}
    \begin{tabular}{@{}lcc@{}}
        \toprule
        \textbf{Model} &
        \textbf{AIME25 Pass@1} &
        \textbf{AIME25 Pass@32} \\
        ToRL-7B        & 27.8 &  30.0   \\
        \rowcolor{blue!10} 
        \hspace{1em} +Ours     & 32.2 &  36.7   \\
        Effective TIR-7B     & 30.0 &  66.7   \\
        \rowcolor{blue!10} 
        \hspace{1em} +Ours     & 36.7 &  70.0  \\
        SimpleTIR-7B        & 28.9 &  53.3   \\
        \rowcolor{blue!10} 
        \hspace{1em} +Ours     & 33.3 &  56.7   \\
       
        \bottomrule

    \end{tabular}
        \caption{\label{tab:ceiling_ability}
     Pass@1 and Pass@32 performances of the proposed ATTC method across various tool integrated reasoning models on AIME 2025.}
\end{table}

\paragraph{Efficiency Experiment}
The results presented in Table~\ref{tab:efficient} demonstrate the token and time consumption of the ATTC approach across various TIR model sizes on five distinct datasets. Overall, ATTC successfully reduces both token and time consumption in the vast majority of scenarios.  Although a few isolated cases show negligible increases, ATTC leads to significant average efficiency gains. Specifically, it results in an average reduction of 5\% to 9\% in token consumption and 9\% to 28\% in time consumption per model. This indicates that ATTC substantially enhances the operational efficiency of TIR models across various scales and tasks.

\subsection{Analysis and Discussion}
\label{sec:Analysis}

\begin{figure}[t]
    \centering
    \includegraphics[width=\columnwidth]{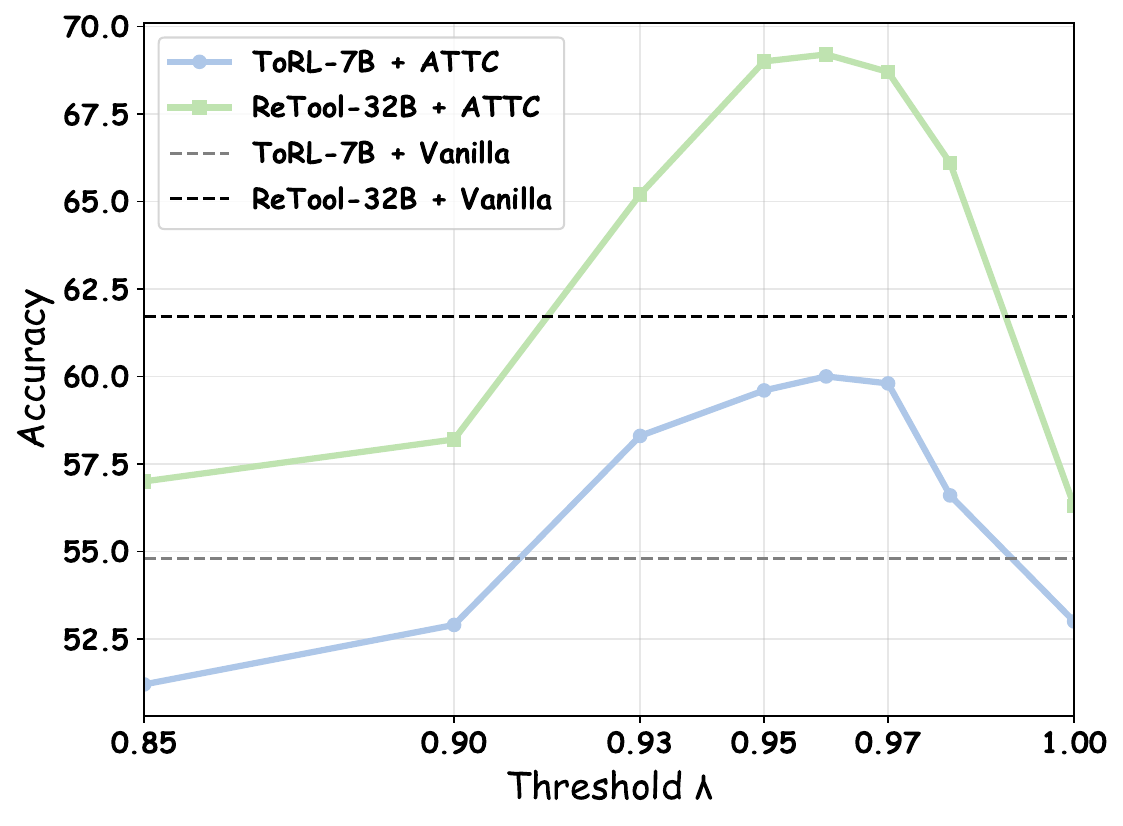}
    \caption{Average performance of ATTC on five datasets under varying settings of the threshold  $\lambda$.}
    \label{fig:threshold}
\end{figure}

\paragraph{Robustness of Threshold $\lambda$}
Figure~\ref{fig:threshold} illustrates the average performance of ATTC on five datasets under varying settings of the threshold  $\lambda$.
The results indicate that when $\lambda$ is set too low, the model completely trusts the tool, leading to a significant decrease in overall accuracy. Conversely, setting $\lambda$ too high causes the model to ignore the tool entirely, also resulting in diminished performance.
Importantly, our method exhibits robustness to $\lambda$ within the range of 0.93 to 0.98, suggesting that fine-grained hyperparameter tuning is not strictly required. The experimental results presented in Table~\ref{tab:main_results} uniformly utilized $\lambda=0.96$. The consistently strong performance across these diverse experiments further validates the generalization ability and robustness of the ATTC approach.

\paragraph{Code Confidence and Reasoning Sufficiency }
\begin{figure}[t]
    \centering
    \includegraphics[width=\columnwidth]{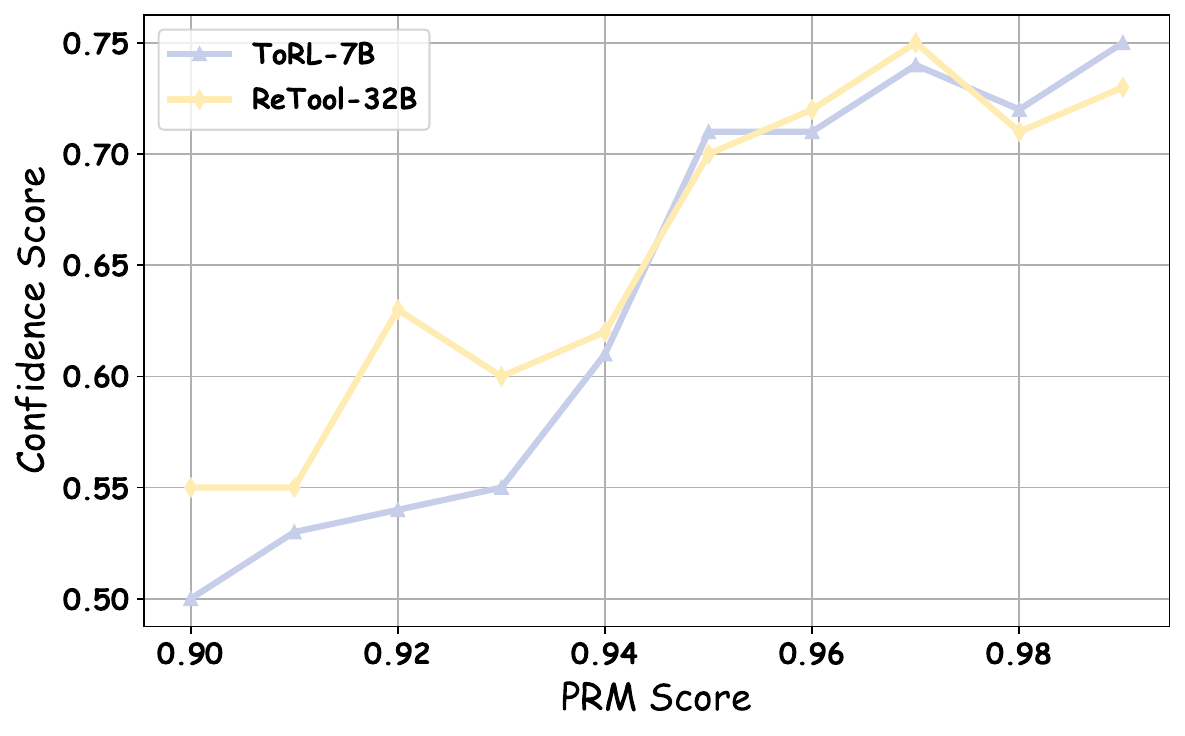}
    \caption{The linear relationship between PRM score and confidence score.}
    \label{fig:prm_confidence}
\end{figure}
To investigate the relationship between a model’s confidence scores assigned to self-generated code blocks and its preceding reasoning process, we employ a process reward model (PRM) to evaluate the sufficiency and rationality of the intermediate reasoning. Specifically, we adopt Universal-PRM-7B~\citep{tan2025auroraautomatedtrainingframeworkuniversal} as the PRM. As shown in Figure~\ref{fig:prm_confidence}, the PRM scores of the model’s preceding reasoning are largely proportional to the confidence scores of the corresponding code blocks. This observation indicates that the confidence scores associated with code blocks can effectively reflect the sufficiency and rationality of the model’s prior reasoning process.

\paragraph{Tool Call Turns}
\begin{figure}[t]
    \centering
    \includegraphics[width=\columnwidth]{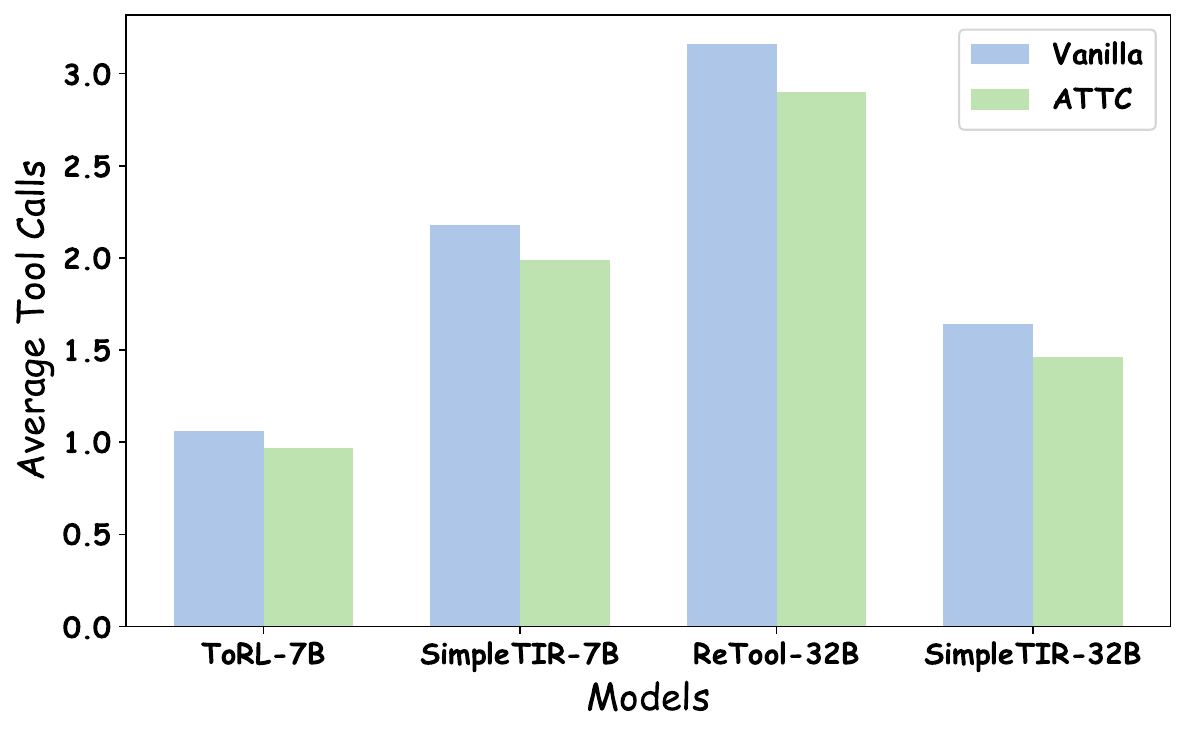}
    \caption{Average number of tool calls per question for four models under the Vanilla and ATTC settings.}
    \label{fig:tool_call}
\end{figure}
Figure~\ref{fig:tool_call} illustrates the average number of tool calls per question for four models under the Vanilla  and ATTC settings. We observe that ATTC does not substantially reduce the overall number of tool calls, its effect is more pronounced for models that tend to perform multiple rounds of tool usage. This behavior aligns with ATTC's design motivation, which focuses on eliminating unnecessary and redundant tool calls while maintaining essential ones rather than aggressively minimizing all tool calls.

\begin{figure}[t]
    \centering
    \includegraphics[width=\columnwidth]{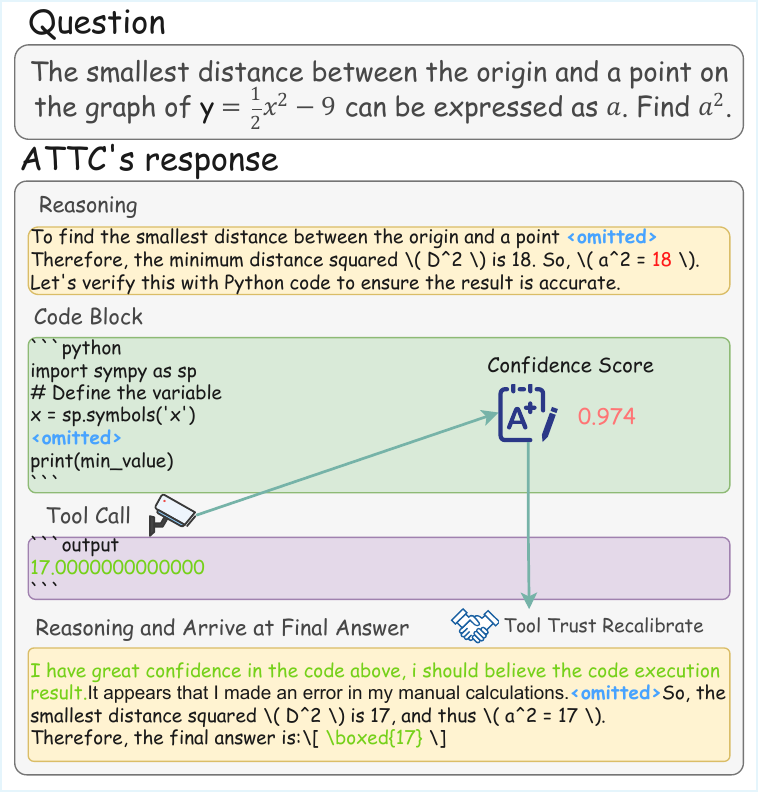}
    \caption{Case of the same problem optimized by ATTC. The model arrive at an incorrect answer of 18 through reasoning, while the tool provide the correct result of 17. The model trust the tool and give an correct answer.}
    \label{fig:case2}
\end{figure}

\paragraph{Tool Ignored Mitigation}
\begin{figure}[t]
    \centering
    \includegraphics[width=\columnwidth]{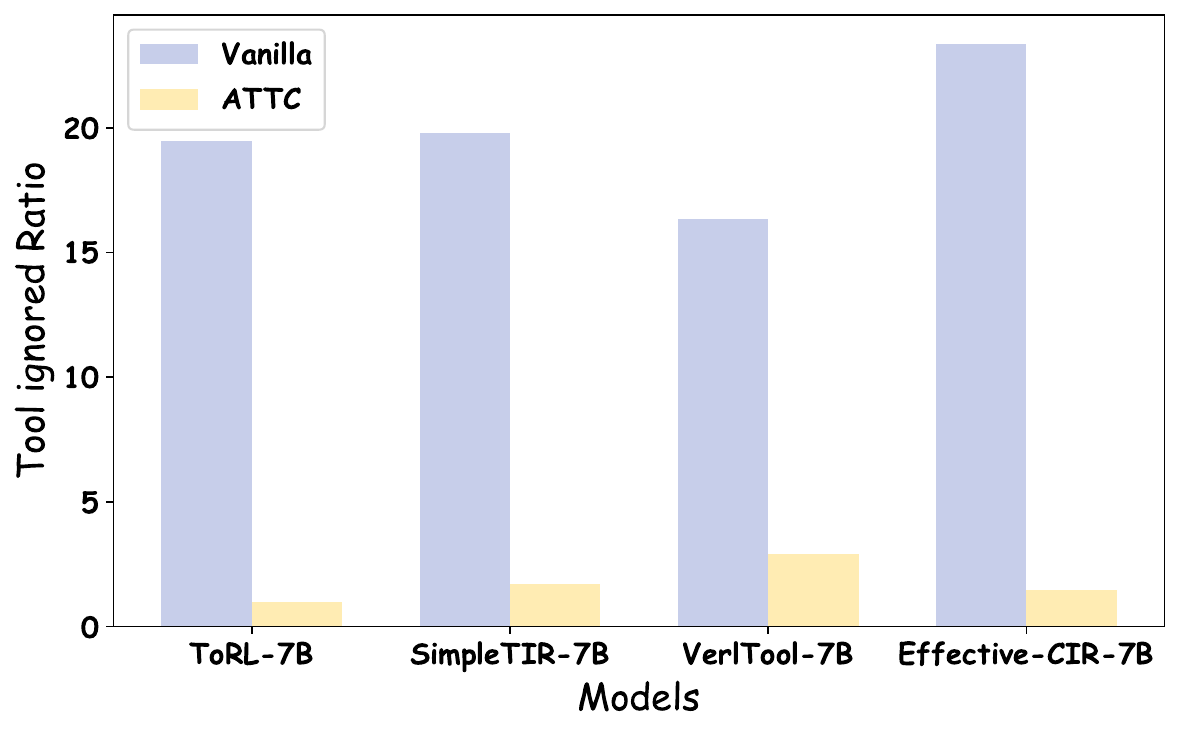}
    \caption{The proportion of ``Tool Ignored''  among false cases for four models across five datasets under the Vanilla and ATTC settings.}
    \label{fig:tool_ignored}
\end{figure}
Figure~\ref{fig:tool_ignored} shows the proportion of ``Tool Ignored''  among false cases for four models across five datasets under the Vanilla and ATTC settings. We observe that ATTC substantially alleviates the ``Tool Ignored'' phenomenon, demonstrating the effectiveness of the proposed method. The specific case of ATTC alleviating ``Tool Ignored'' is shown in Figure~\ref{fig:case2}.
Although ATTC greatly decreases the number of issues, there are still some cases where ``Tool Ignored'' appears. This is due to the inherent tendency of large language models to hallucinate, which makes it difficult for them to completely follow the guidance provided by ATTC. Addressing this limitation may require further investigation in future research.

\section{Related work}

\subsection{Tool Integrated Reasoning}
Tool-Integrated reasoning (TIR)~\citep{gou2024toratoolintegratedreasoningagent, wang2023mathcoderseamlesscodeintegration, liao2024mariomathreasoningcode}   is an advanced paradigm that seamlessly incorporates the call and execution of external tools into the reasoning process. By leveraging external tools such as code executors and search engines, TIR enables LLMs to transcend the inherent limitations of purely reasoning\citep{lin2025understandingtoolintegratedreasoning}.
Early works in  TIR primarily relied on prompt engineering~\citep{wang2025selfdcreasonactself, yuan2024craftcustomizingllmscreating, qian2024investigateconsolidateexploitgeneralstrategyintertask,chen2023programthoughtspromptingdisentangling, yang2024bufferthoughtsthoughtaugmentedreasoning} to guide LLMs in tool call. However, these approaches were heavily rely on meticulously crafted prompts, which limited their scalability and generalizability. 
Some later works used supervised finetuning ~\citep{chen2025advancingtoolaugmentedlargelanguage, qian2025smartselfawareagenttool, gou2024toratoolintegratedreasoningagent, 
liao2024mariomathreasoningcode, schick2023toolformerlanguagemodelsteach} to solidify the behavior pattern of the model actively calling tools in reasoning. Qwen2.5-Math-Instruct ~\citep{yang2024qwen25mathtechnicalreportmathematical} , ReAct~\citep{yao2023reactsynergizingreasoningacting} and MathCoder~\citep{wang2023mathcoderseamlesscodeintegration} advances this paradigm by training models on specialized datasets enriched with tool call demonstrations.
Through finetuning, the model transitions from passive response to proactive tool call during the reasoning process, where it can autonomously trigger external tools and seamlessly integrate the returned execution results to sustain the reasoning trajectory.

\subsection{RL for Tool Integrated Reasoning}
SFT-based methods exhibit inherent limitations, as they constrain models to strictly adhere to the tool usage patterns present in the training data distribution, these models often fail to develop adaptive strategies based on the difficulty of the task.
Several recent works focus on applying reinforcement learning\citep{shao2024deepseekmathpushinglimitsmathematical, deepseekai2025deepseekr1incentivizingreasoningcapability, yu2025dapoopensourcellmreinforcement, liu2025understandingr1zeroliketrainingcritical, zeng2025simplerlzooinvestigatingtamingzero} to address these issues.
ReTool~\citep{feng2025retoolreinforcementlearningstrategic} proposes an automated RL paradigm that allows policy rollouts with multi-turn code execution.
ToRL~\citep{li2025torlscalingtoolintegratedrl} enables models to discover optimal strategies for tool utilization via unrestricted exploration.
VerlTool~\citep{jiang2025verltoolholisticagenticreinforcement} introduces a unified and modular framework to address multiturn tool interactions.
Effective CIR~\citep{bai2025effectivecodeintegratedreasoning} develops enhanced training strategies that balance exploration and stability, progressively
building tool-use capabilities while improving reasoning performance.
SimpleTIR~\citep{xue2025simpletirendtoendreinforcementlearning} stabilizes multi-turn TIR training by filtering out trajectories containing turns that yield neither a code block nor a final answer. 




\section{Conclusion}

In this work, we find some shortcomes in the existing open-source TIR models. There is a common contradiction between the reasoning of model and the results of external tools, and the model tends to believe in its own reasoning. We also discover and define ``Tool Ignored'', which represents the situation where the correct results of the tool are ignored by the model, resulting in incorrect answers. We propose a new framework ATTC, and the experimental results from various TIR models and across multiple datasets demonstrate that ATTC effectively reduces the ``Tool Ignored'' issue, resulting in a performance increase of 4.1\% to 7.5\%.


\section*{Limitations}

Although ATTC has generally improved the performance of TIR models, we identify some possible limitations as follows:
\begin{itemize}[leftmargin=*]
    \setlength{\itemsep}{0pt}
    \setlength{\parskip}{0pt}
    \item Due to limitations in computing resources, we only conducted experiments on open-source TIR models with a size less than 32B.
    \item This work focuses exclusively on the application of TIR in scenarios where code executors are utilized as the tool. We have not yet extended our method to scenarios where search engines are utilized as the tool, which we leave for future exploration.
    \item Currently, optimization is only being conducted during the reasoning stage. In the future, we will consider addressing the problem at its root during training.
\end{itemize}

\section*{Acknowledgement}
We want to thank all the anonymous reviewers for their valuable comments. This work was supported by the National Science Foundation of China (NSFC No. 62576232), Key Laboratory of Data Intelligence and Advanced Computing, Soochow University.

\bibliography{main}

\clearpage
\appendix
\section{The Use of Large Language Models}
\label{sec:appendixA}
We use large language models to assist in analyzing the reasoning trajectories of a large number of TIR models.
A large language model was used in writing this manuscript to assist with proofreading and improve the clarity of the text. All knowledge content, including ideas, analysis, and conclusions, is the author's work.

\section{Prompt Template}
\label{sec:appendixB}
The prompt template used to detect the conflict rate between reasoning and tools is shown in the Figure~\ref{fig:contradiction_prompt}.
The prompt template for detecting  ``Tool Ignored'' is shown in the Figure~\ref{fig:tool_ignored_prompt}.
The specific prompt template for different TIR models are shown in Figures~\ref{fig:ToRL_prompt} to ~\ref{fig:DemyAgent_prompt}.

\begin{figure*}[t] 
\centering
\begin{tcolorbox}[colback=white, colframe=white!60!black, coltitle=black, colbacktitle=purple!20!white, 
title=Prompt Template For Detecting The Contradiction Ratio Between Reasoning And Tools,
fonttitle=\bfseries, enhanced, 
attach boxed title to top center={yshift=-2mm},
width=\textwidth] 
\begin{verbatim}
I will give you a jsonl formatted data, which is a case where the model answered.
"question" is the question answered by the model, "answer" is the standard answer,
"solution" is the standard problem-solving process,
"code" is the content of the model's answer, and
"pred" is the final answer of the model. 
You need to analyze the model's answer content based on this information. 

If there is a contradiction between the inference of the model and the execution 
result of Python code (after "```output") ,the model chooses to believe in the 
inference, output "\\boxed{0}".
If there is a contradiction between the inference of the model and the execution
result of Python code(after "```output"), the model chooses to believe in the code
execution result,output "\\boxed{1}".If neither of these situations exist,
output "\\boxed{2}".Your answer is limited to these three types and must be included
in "\\boxed{}, don't output extra text".
The cases that need to be analyzed are as follows:\n
\end{verbatim}
\end{tcolorbox}
\caption{Contradiction Ratio Prompt Template}
\label{fig:contradiction_prompt}
\end{figure*}

\begin{figure*}[t] 
\centering
\begin{tcolorbox}[colback=white, colframe=white!60!black, coltitle=black, colbacktitle=purple!20!white, 
title=Prompt Template For Detecting ``Tool Ignored'',
fonttitle=\bfseries, enhanced, 
attach boxed title to top center={yshift=-2mm},
width=\textwidth] 
\begin{verbatim}
I will give you a jsonl formatted data, which is a case where the model answered.
"question" is the question answered by the model, "answer" is the standard answer,
"solution" is the standard problem-solving process,
"code" is the content of the model's answer, and
"pred" is the final answer of the model. 
You need to analyze the model's answer content based on this information. 

If there is no Python code segment(after "```python") in the model's answer, output
"\\boxed{2}". If the return result of the Python code (after "```output") is correct
(including correct answer and correct thinking), output "\\boxed{1}". Otherwise,
output "\\boxed{0}". Your answer is limited to these three types and must be
included in "\\boxed{}, don't output extra text".
The cases that need to be analyzed are as follows:\n
\end{verbatim}
\end{tcolorbox}
\caption{``Tool Ignored'' Prompt Template}
\label{fig:tool_ignored_prompt}
\end{figure*}

\section{Experimental Set}
\label{sec:appendixC}
The specific temperature and top\_p settings for different TIR models are shown in the table~\ref{tab:tem_top}

\begin{table}[H]
    \centering
    \footnotesize
    \setlength{\tabcolsep}{5.5pt}     
    \renewcommand{\arraystretch}{1.15} 
    \vspace{0.2cm}
    \begin{tabular}{@{}lcc@{}}
        \toprule
        \textbf{Model} &
        \textbf{Temperature} &
        \textbf{Top\_p}   \\
        \midrule
        ToRL-7B        & 0 &  1   \\
        Effective TIR-7B          & 0.6 &  0.95   \\
        VerlTool-7B         & 0 &  1  \\
        SimpleTIR-7B\&32B        & 1.0 &  0.7  \\
        ReTool-32B\&4B   & 1.0 &  0.7  \\
        DemyAgent-4B    & 1.0 &  0.6  \\

        \bottomrule
    \end{tabular}
    \caption{\label{tab:tem_top}
    Temperature and top\_p of different TIR models.}
\end{table}

\section{Method Details}
\label{sec:appendixD}
During the Tool Trust Recalibration process, when $\mathcal{C} \ge \lambda$, ATTC guides TIR to trust the tool's results and continue reasoning by inserting special prompts into the original reasoning trajectory. 
On the contrary, when $\mathcal{C} < \lambda$, ATTC guides the TIR model to consider rewriting the code or rethinking the reasoning path based on the tool results by inserting additional prompts. The specific prompt is shown in the Table~\ref{tab:method_details}.

\begin{figure*}[t] 
\centering
\begin{tcolorbox}[colback=white, colframe=white!60!black, coltitle=black, colbacktitle=blue!20!white, 
title=Prompt Template For ToRL-7B\&VerlTool-7B,
fonttitle=\bfseries, enhanced, 
attach boxed title to top center={yshift=-2mm},
width=\textwidth] 
\begin{verbatim}
A conversation between User and Assistant. The user asks a question, and 
the Assistant solves it.
User:Please integrate natural language reasoning with programs to solve the problem 
above, and put your final answer within \boxed{}.
{Question}
Assistant:
\end{verbatim}
\end{tcolorbox}
\caption{ToRL-7B\&VerlTool-7B's Prompt Template}
\label{fig:ToRL_prompt}
\end{figure*}

\begin{figure*}[t] 
\centering
\begin{tcolorbox}[colback=white, colframe=white!60!black, coltitle=black, colbacktitle=blue!20!white, 
title=Prompt Template For Effective TIR-7B,
fonttitle=\bfseries, enhanced, 
attach boxed title to top center={yshift=-2mm},
width=\textwidth] 
\begin{verbatim}
Please solve the following problem step by step. During your reasoning process, 
if needed, you can choose to write python code to enhance your reasoning.
The code executor will run your code and provide the execution results back to you
to support your reasoning process.
Please put the final answer within \boxed{}.
Question:
{Question}
\end{verbatim}
\end{tcolorbox}
\caption{Effective TIR-7B 's Prompt Template}
\label{fig:EffecTIR_prompt}
\end{figure*}


\begin{figure*}[t] 
\centering
\begin{tcolorbox}[colback=white, colframe=white!60!black, coltitle=black, colbacktitle=blue!20!white, 
title=Prompt Template For SimpleTIR-7B\&32B,
fonttitle=\bfseries, enhanced, 
attach boxed title to top center={yshift=-2mm},
width=\textwidth] 
\begin{verbatim}
Solve the following problem step by step. You now have the ability to
selectively write executable Python code to enhance your reasoning
process. The Python code will be executed by an external sandbox,
and the output (after “Code execution result: ”) is returned to aid your
reasoning and help you arrive at the final answer. The Python code
should be complete scripts, including necessary imports.

Code Format:
Each code snippet is wrapped between \\n```python \\n```.\\n You need to use
`print()` to output intermediate results.The code execution result is only 
auxiliary, you can choose to believe or not believe it.

Answer Format:
You should use \\boxed to return your final answer. The last part of your response
should be: \boxed{'The final answer goes here.'}

User Question:
{Question}
\end{verbatim}
\end{tcolorbox}
\caption{SimpleTIR-7B\&32B 's Prompt Template}
\label{fig:SimpleTIR_prompt}
\end{figure*}

\begin{figure*}[t] 
\centering
\begin{tcolorbox}[colback=white, colframe=white!60!black, coltitle=black, colbacktitle=blue!20!white, 
title=Prompt Template For ReTool-32B\&4B ,
fonttitle=\bfseries, enhanced, 
attach boxed title to top center={yshift=-2mm},
width=\textwidth] 
\begin{verbatim}
Solve the following problem step by step. You now have the ability to selectively
write executable Python code to enhance your reasoning process. The Python code
will be executed by an external sandbox, and the output (wrapped in <interpreter>
output</interpreter>) can be returned to aid your reasoning and help you arrive at 
the final answer. 
The Python code should be complete scripts, including necessary imports.

Code Format:
Each code snippet is wrapped with
<code>
```python
code snippet
```
</code>

Answer Format:
The last part of your response should be:
<answer>\boxed{'The final answer goes here.'}</answer>

User Question:
{Question}
Assistant:
\end{verbatim}
\end{tcolorbox}
\caption{ReTool-32B\&4B's Prompt Template}
\label{fig:ReTool_prompt}
\end{figure*}

\begin{figure*}[t] 
\centering
\begin{tcolorbox}[colback=white, colframe=white!60!black, coltitle=black, colbacktitle=blue!20!white, 
title=Prompt Template For DemyAgent-4B,
fonttitle=\bfseries, enhanced, 
attach boxed title to top center={yshift=-2mm},
width=\textwidth] 
\begin{verbatim}

Analyze and solve the following [math/science domain] problem step by step.
Problem: {Question}
Hint: The tool could be used for more precise and efficient calculations and could
help you to verify your result before you reach the final answer.

Note: You should first analyze the problem and form a high-level solution strategy, 
then utilize the tools to help you solve the problem.

Answer Format: Do not put units of the final answer inside \boxed{}. The content of
\boxed{} should be the numerical value of the final answer only, without any units.
Remember once you make sure the current answer is your final answer, do not call the
tools again and directly output the final answer in the following text format, 
the answer format must be: \boxed{’The final answer goes here.’}.

\end{verbatim}
\end{tcolorbox}
\caption{DemyAgent-4B\&Qwen3-4B's Prompt Template}
\label{fig:DemyAgent_prompt}
\end{figure*}



\begin{table}[H]
    \centering
    \footnotesize
    \setlength{\tabcolsep}{5pt}
    \renewcommand{\arraystretch}{1.15}
    \begin{tabular}{@{}lp{5cm}@{}}
        \toprule
        \textbf{Control Signal} & \textbf{Prompt} \\
        \midrule
        Trust   & I have great confidence in the code above, i should believe the code execution result. \\
        Rethink & I have no confidence in the code above, i should consider optimizing it or reflecting on whether to change my reasoning path. \\
        \bottomrule
    \end{tabular}
    \caption{Method Details.}
    \label{tab:method_details}
\end{table}

\section{ Case Studies}
\label{sec:appendixE}
\begin{figure*}[t]
    \centering
    \includegraphics[width=\textwidth]{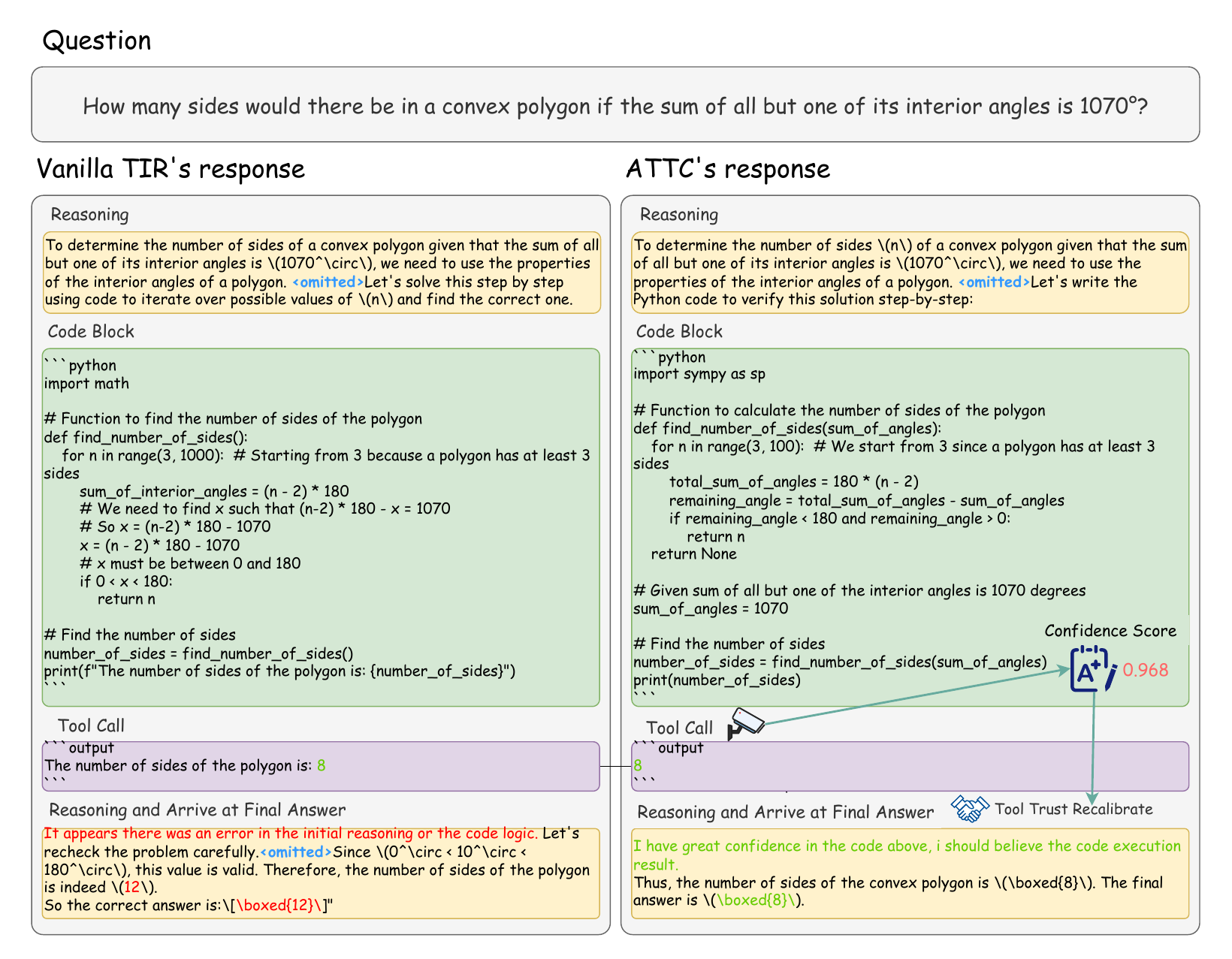}
    \caption{A case of the comparison between the Vanilla TIR response and ATTC's response.}
    \label{fig:case3}
\end{figure*}

Figure~\ref{fig:case3} shows a case of the comparison between the Vanilla TIR response and ATTC's response. In vanilla TIR, the model ignores the correct answer provided by the tool and answers incorrectly. In contrast, the ATTC framework guides the model to trust the tool's results and provide accurate answers.

\end{document}